\documentclass[letterpaper, 10 pt, journal]{IEEEtran}  
\IEEEoverridecommandlockouts

\usepackage{amsthm}
\usepackage{algorithm, algorithmicx, algpseudocode}
\usepackage{amsmath,amssymb,enumerate}
\usepackage{mathrsfs}
\usepackage{times}
\usepackage{multicol}
\usepackage{amsfonts}
\usepackage{textcomp}
\usepackage{balance}
\usepackage{multirow}
\usepackage{booktabs}
\usepackage{dblfloatfix} 
\usepackage{bbold}
\usepackage{makecell}
\usepackage{hhline} % double-line in a table
\usepackage{float} % Places the float at precisely the location in the code
\usepackage[mathscr]{euscript}
\usepackage[sort&compress,numbers]{natbib}

\usepackage{comment}
\usepackage{xparse}
\usepackage{lipsum}
\usepackage{soul} 
\usepackage{balance}
\usepackage{dsfont} % indicator function
\usepackage{setspace} % double spacing
\usepackage{accents} % widetilde

\usepackage{graphicx}
\usepackage{fixltx2e}

\usepackage[T1]{fontenc}
\usepackage[usenames,dvipsnames,svgnames,table, xcdraw]{xcolor}
\usepackage[caption=false,font=footnotesize]{subfig}
\usepackage[utf8]{inputenc}
\usepackage{hyperref}
\hypersetup{
  colorlinks=true,pdfpagemode=UseNone,citecolor=black,linkcolor=black,urlcolor=BrickRed
}

\usepackage{caption}
\graphicspath{
    {fig/}
}

\newcommand{\transpose}{\mathsf{T}}

\newcommand{\m}{\mathop{\mathrm{m}}}

\makeatletter
\newcommand{\mathleft}{\@fleqntrue\@mathmargin0pt}
\newcommand{\mathcenter}{\@fleqnfalse}
\makeatother

\usepackage{epsfig} % for postscript graphics files
\usepackage{bm}

\title{Fusing Convolutional Neural Network and Geometric Constraint for Image-based Indoor Localization }%

\author{Jingwei Song, Mitesh Patel, and Maani Ghaffari%
	\thanks{J. Song and M. Ghaffari are with the University of Michigan, Ann Arbor, MI 48109, USA. \texttt{\{jingweso,maanigj\}@umich.edu}}%
	\thanks{M. Patel is with Nvidia Corporation, Santa Clara, CA - 95051, USA. \texttt{\{miteshp\}@nvidia.com}}%
}

\begin{document}

	\maketitle
% 	\thispagestyle{empty}
% 	\pagestyle{empty}

	%%%%%%%%%%%%%%%%%%%%%%%%%%%%%%%%%%%%%%%%%%%%%%%%%%%%%%%%%%%%%%%%%%%%%%%%%%%%%%%%
	\begin{abstract}
	    This paper proposes a new image-based localization framework that explicitly localizes the camera/robot by fusing Convolutional Neural Network (CNN) and sequential images' geometric constraints. The camera is localized using a single or few observed images and training images with 6-degree-of-freedom pose labels. A Siamese network structure is adopted to train an image descriptor network, and the visually similar candidate image in the training set is retrieved to localize the testing image geometrically. Meanwhile, a probabilistic motion model predicts the pose based on a constant velocity assumption. The two estimated poses are finally fused using their uncertainties to yield an accurate pose prediction. This method leverages the geometric uncertainty and is applicable in indoor scenarios predominated by diffuse illumination. Experiments on simulation and real data sets demonstrate the efficiency of our proposed method. The results further show that combining the CNN-based framework with geometric constraint achieves better accuracy when compared with CNN-only methods, especially when the training data size is small. 
	\end{abstract}
	
\begin{IEEEkeywords}
Vision-Based Navigation, Localization, Probability and Statistical Methods.
\end{IEEEkeywords}

\IEEEpeerreviewmaketitle

	%%%%%%%%%%%%%%%%%%%%%%%%%%%%%%%%%%%%%%%%%%%%%%%%%%%%%%%%%%%%%%%%%%%%%%%%%%%%%%%%

	\section{Introduction}
	Image-based localization refers to retrieving 6 Degree-of-Freedom (DoF) pose of the camera (monocular in most cases) using the obtained single or few images (testing) and previous pose labeled images (training) \cite{patel2018contextualnet}. Typical applications include pedestrian localization with one image taken by mobile phone or localization of service robot in indoor with low-cost camera.
	
	Contrary to prior-free monocular Simultaneous Localization And Mapping (SLAM), image-based localization can align the observed single/few images to the training images' coordinate and recover the pose globally. In monocular SLAM, the system can only recover the camera pose locally (relative to the starting pose) using a sequence of images without the correct scale. The scale and pose in monocular SLAM suffer from drift \cite{mur2017orb}. Table~\ref{Table_diff_SLAM} summarizes the differences. Since image-based localization works with consumer-level monocular cameras, it provides a fast and low-cost solution for a wide variety of applications.\par 
	
	\begin{figure}[t]    
		\centering
		\includegraphics[width=0.99\linewidth]{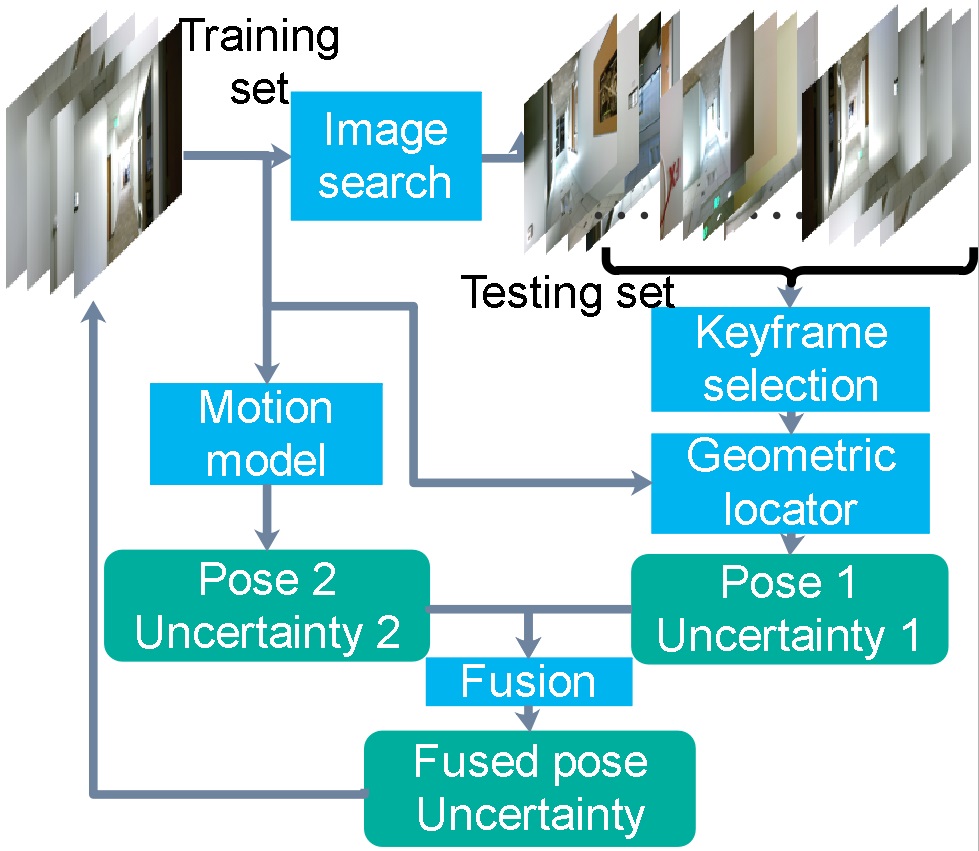}            
		\caption{The proposed method contains five major parts colored in blue: The similarity-based image locator, keyframe selection module, the geometric locator, the motion model, and the fusion module. The geometric locator and motion model are two parallel processes, and the fusion module combines the two predictions. The images are processed in a non sequential manner.}
		\label{fig:Flowchart}
	\end{figure}
	
	\begin{table}[]
	\scriptsize
  \centering
		\caption{The table shows the difference between monocular image-based localization and monocular SLAM.}
    \begin{tabular}{l|p{2.85cm}<{\centering}|p{2.85cm}<{\centering}}
                     & Monocular SLAM & Image-based localization \\ \midrule
    Prior data       & N/A            & Images and pose labels   \\
    Scale            & No             & Yes                      \\
    Pose coordinate  & Local          & Global                   \\
    Number of images & Sequential     & Single or few            \\ \bottomrule
    \end{tabular}
    \label{Table_diff_SLAM}
    \end{table}

	Previous researches aligned the obtained image with the viewed scenery in the training set based on the visual similarity (observed from a similar viewing angle). The similarity is encoded in feature extractors like Speeded Up Robust Features (SURF) or Oriented FAST and Rotated BRIEF (ORB). Later, PoseNet~\cite{kendall2015posenet}, and SCORE forest~\cite{shotton2013scene} showed the machine learning method, especially the deep Convolutional Neural Network (CNN), can map the images to its 6 DoF poses. The trained deep CNN model was used to infer the pose of the testing image directly. The follow up work developed different variants of PoseNet to enhance the performance of image-to-pose deep CNN models~\cite{wang2019atloc,weinzaepfel2019visual,valada2018deep,patel2018contextualnet,radwan2018vlocnet++,tian20203d}. \par

    To estimate the uncertainties for the image-to-pose deep CNN model, \citet{kendall2016modelling} adopted a Bayesian deep CNN framework to quantify the estimated pose's uncertainty.	It addressed uncertainty based on the training data set coverage, as it maintained the assumption that the ``model is more uncertain about images which are dissimilar to the training examples'' \cite{kendall2016modelling}. The epistemic uncertainty measures the output in a data-fitting manner. Similar studies can be found in \cite{zhao2019generative,costante2020uncertainty, yang2020d3vo}. In the Bayesian deep CNN network, the pose uncertainty refers to the ambiguity caused by the unequaled distributions between the training and testing samples, distinguishing itself from the traditional geometric uncertainty. Unlike the Bayesian deep CNN model, \citet{liu2018deep} inferred the covariance of the measurement from raw sensory data directly. In contrast to the data-driven methods, the model-driven aleatoric uncertainty~\cite{hartley2003multiple,vega2013cello,ozog2013importance,polic2017camera,briskin2017estimating} addresses the geometric uncertainty. These geometric works followed the classical pinhole camera model and obtained the 6 DoF pose uncertainty propagated from the uncertainty in 2D tracking, measured as the 2D re-projection residuals sum.\par

	In general, deep CNN-based image-to-pose approaches are only efficient if images in the training and testing set are visually similar. Deep CNN-based methods cannot retrieve good results if the images are collected from very different perspectives. Moreover, the uncertainties in the deep CNN framework measure the similarity between the training and testing images. The obtained pose uncertainties from the Bayesian network have only valid statistical meaning instead of geometric interpretations. Our work exploits the pinhole camera projection model and is less dependent on the size of the training data. Moreover, benefiting from the geometric uncertainty quantification, we use a constant velocity camera motion model to incorporate the information of consecutive images to gain robustness. By introducing the motion model, we overcome the issues caused by blurry images, which randomly yields inaccurate far-out predictions. The pose predictions provided by the camera projection model and motion model are further fused based on the uncertainties.

	The contributions of this work are as follows.
	\begin{enumerate}
		\item A novel image-based localization framework that explicitly localizes the robot by fusing deep CNN and sequential images' geometric constraints.
		\item The predictions from the geometric locator and the motion model are fused by incorporating their respective uncertainties during pose fusion.
		\item Our proposed framework requires relatively small training data set and can be easily deployed for indoor localization tasks.
	\end{enumerate}

	\section{Related Work}

	Image-based localization can be categorized as single monocular and sequential monocular image-based localization.\par 
	
 	Early researches used the monocular image captured by the consumer-level camera for image-based localization. The prior-free Bag of Words (BoW) algorithm~\cite{galvez2012bags} obtains the image descriptors to characterize the image appearance. Due to its efficiency, BoW has been widely applied in SLAM systems for loop closure detection~\cite{galvez2012bags}. Following BoW, other methods used SIFT probability~\cite{cummins2008fab}, deep CNN framework~\cite{sunderhauf2015performance} and contextual re-weight network~\cite{kim2017learned} to generate image descriptors. All these methods only provide rough camera localization and not the precise 6 DoF pose.
	
	PoseNet~\cite{kendall2015posenet} (monocular) and SCORE forest~\cite{shotton2013scene} (RGB-D) are learning-based approaches to map from image-to-6 DoF pose. The learning-based approaches bridge the gap between the learning and the geometric domain. 
	\citet{wang2019atloc} introduced an attention model for better feature extraction. \citet{weinzaepfel2019visual} claimed the benefits of an intermediate objects-of-interest extraction and matching. \citet{tian20203d} trained a deep CNN model with better efficiency by addressing the 3D scene geometry-aware constraints in consecutive images. The idea proposed by~\citet{tian20203d} is similar to our approach where geometry-aware constraints are used via an implicit encoding in the training process. However, we explicitly convert the image-to-pose CNN into a hybrid of image-to-image matching and geometric localization. We also exploit the inter-image relationship by explicitly using a motion model constraint.\par

	Considering that short-term consecutive images may benefit the robustness and accuracy of the image-to-pose fitting, further studies improved the deep CNN-based framework with sequential images. \citet{patel2018contextualnet} proposed a Long Short Term Memory (LSTM) network termed as ContextualNet. The information within sequential images is exploited to enhance the performance of the deep CNN network. VLocNet~\cite{valada2018deep} and VLocNet$++$ \cite{radwan2018vlocnet++} integrated multiple deep CNNs to predict 6 DoF pose, visual odometry, and semantic map segmentation simultaneously. This multi-task deep CNN strategy significantly enhanced the performance of the pose prediction. \citet{brahmbhatt2018geometry, yang2020d3vo} integrated conventional SLAM technique by feeding the redundant pose predictions of multiple images to the pose graph or Bundle Adjustment (BA) optimizations.  \par

	\section{Notation}

	Denote $(\cdot)^\vee:\mathfrak{se}(3) \to \mathbb{R}^6$ as an isomorphism that converts an element of Lie algebra $\mathfrak{se}(3)$ to a 6-vector (i.e., twist). Adversely, $(\cdot)^\wedge:\mathbb{R}^6 \to \mathfrak{se}(3)$. For the testing image $i$, the goal of this work is to estimate the camera's pose $\mathbf{T}^F_i \in \mathrm{SE}(3)$. Additionally, the associated covariance matrix is denoted ${\mathbf{C}}^F_{i} \in \mathbb{R}^{6\times6}$. We define $\operatorname{d}_{\mathfrak{se}(3)}(\mathbf{T}_1,\mathbf{T}_2)$ as a distance between two poses $\mathbf{T}_1, \mathbf{T}_2 \in \mathrm{SE}(3)$ as

	\begin{equation*}
	\operatorname{d}_{\mathfrak{se}(3)}(\mathbf{T}_1,\mathbf{T}_2) := \lVert \operatorname{log}(\mathbf{T}^{-1}_1\mathbf{T}_2)^{\vee} \rVert_2,
	\end{equation*}

	\noindent where $\operatorname{log}(\cdot)$ is the matrix (Lie) logarithm map. The misalignment angle between two orientations $\mathbf{R}_1, \mathbf{R}_2 \in \mathrm{SO}(3)$ is

	\begin{equation}
	\label{Eq_radius_angle}
	\Theta(\mathbf{R}_1^\transpose \mathbf{R}_2) = \operatorname{acos}((\operatorname{Tr}(\mathbf{R}_1^\transpose \mathbf{R}_2)-1)/2),
	\end{equation}
	\noindent where $\operatorname{Tr}(\cdot)$ yields the trace of the input matrix. Denote $\mathbf{T}|_r\in \mathrm{SO}(3)$ and $\mathbf{T}|_p\in \mathbb{R}^3$ as the orientation and position from the pose $\mathbf{T}\in \mathrm{SE}(3)$.\par

	\section{Proposed Methodology}
	Fig.~\ref{fig:Flowchart} illustrates the proposed framework, which consists of five modules: (a) Aligning the single testing image to the most visually similar image in the training set. (b) Selecting co-visible keyframes to the visually similar image in the training set. (c) The testing image is localized using the selected co-visible images using geometric features. (d) The camera pose of the testing image is retrieved using a motion model based on the constant velocity assumption to account for the sequential predictions. (e) The poses predicted from the geometric locator and the motion model are fused, addressing both uncertainties. If the image has no sequential prior predictions, only the single image locator is used in the first few starting frames.\par

	\subsection{Image descriptor network}
	\label{section_classification_Siamese}
	
	The image descriptor network aims at estimating the camera pose by aligning the most visually similar image in the training set with the testing image, using the trained deep CNN model. We modify the end-to-end image-to-pose regression network PoseNet \cite{kendall2015posenet} as an image-to-image similarity searching for following reasons. First, \citet{kendall2016modelling} pointed out that the uncertainty of the PoseNet is closely related to the similarity of the training and testing images. The Siamese network structure can be explicitly used to compare the image similarities to exploit the information better. Second, \citet{zagoruyko2015learning} demonstrates that the Siamese network with deep CNN model is efficient in measuring image similarity with limited training data. Lastly, BA addresses the inherent geometric constraint and provides an error propagation formulation mathematically derived from the pinhole camera model.\par
	
\begin{figure}[t]
		\centering
		\includegraphics[width=0.99\columnwidth]{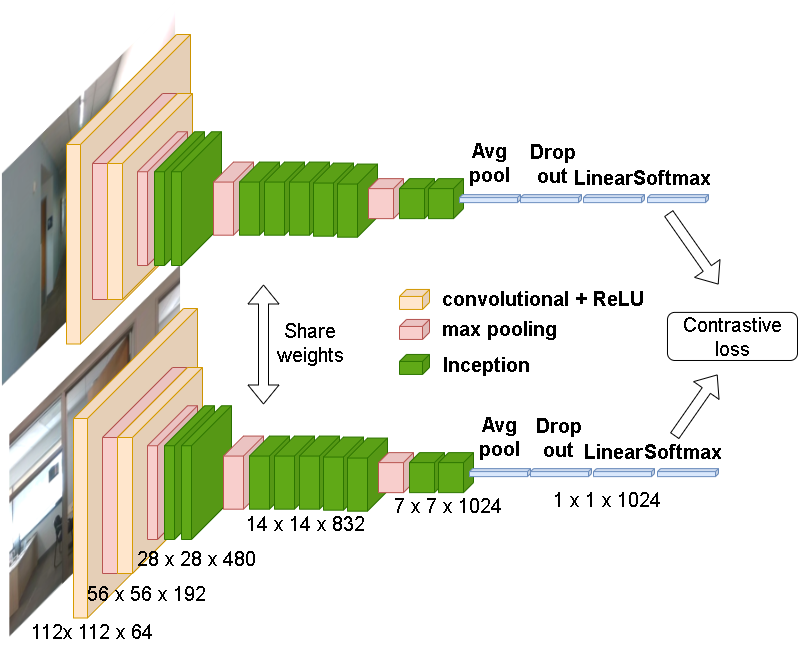}
		\caption{Illustrated is the structure of the Siamese network used for training the image descriptor network, which is a modified version of GoogLeNet.}
		\label{fig:Siamese_network}
	\end{figure}

	Fig. \ref{fig:Siamese_network} shows the structure of the Siamese network structure. Among the off-the-shelf methods like  \cite{GalvezTRO12,schmitt2017openxbow}, or deep CNN-based method \cite{wang2014learning,zagoruyko2015learning}, GoogLeNet~\cite{szegedy2015going} is adopted as the base network since it works well in PoseNet \cite{kendall2015posenet} in extracting image descriptors. The twin GoogLeNet models share the same weights. All input images are regularized to grayscale images with $224\times 224$. Each twin network takes one of the two gray images. The branch's outputs are concatenated and fed into a fully connected layer to generate a one-dimensional image descriptor. In the training process, arbitrary thresholds of the pose are chosen to label the similarity of the image. We used contrastive loss function \cite{hadsell2006dimensionality} to measure the similarity between the encoded descriptors \eqref{Eq_constrastive_loss}. $\mathbf{s}_1$ and $\mathbf{s}_2$ are vectors yielded from the image descriptor networks. Before training, we build a K-D tree on the descriptors of the training images, which is further used to find the most similar training image using the image descriptors. In~\eqref{Eq_constrastive_loss}, $\mathrm{m}$ is the margin. 
	\begin{equation}
	\label{Eq_constrastive_loss}
	\begin{aligned} 
	L\left(\mathbf{s}_1, \mathbf{s}_2\right)=(1-\mathbf{s}_1) \frac{1}{2}\left(\mathbf{s}_2\right)^{2}+(\mathbf{s}_1) \frac{1}{2}\left\{\max \left(0, \mathrm{m}-\mathbf{s}_2\right)\right\}^{2}. \end{aligned}
	\end{equation}
	
	\subsection{Co-visible keyframes selection}
	\label{section_neighbor_select}

	After the testing image is aligned with the optimal training image, the optimal training image's co-visible keyframes are selected following two principles: (I) Every two keyframes should have a suitable baseline to allow enough parallax and overlap for triangulation. (II) The orientational difference between the two images is trivial so that there is enough overlap for triangulation. (I) ensures large enough parallax while (II) is to avoid too small overlap within the selected images. We follow \cite{ondruvska2015mobilefusion} and modify the keyframe selection strategy by calculating the score. For two arbitrary keyframes with camera poses $\mathbf{T}_1$ and $\mathbf{T}_2$, the score is defined as 
	\begin{align*}
	\operatorname{s}&(\mathbf{T}_1\!,\!\mathbf{T}_2\!) \!=\!
	\exp\! \left(\!-\frac{\left(\lVert\mathbf{p}_1^2- \mathbf{p}_2^2\lVert_2\!-\!d_{m}\right)^{2}}{\ell^{2}}\! \right)\! \cdot 
	\!	\max\! \left(\frac{\alpha_{m}}{\Theta\left(\!\mathbf{R}_1^\top\!\mathbf{R}_2\!\right)},\! 1\!\right),
	\end{align*}

	\noindent where $\mathbf{T}_1$ is composed of orientation $\mathbf{R}_1$ and position $\mathbf{p}_1$ and $\mathbf{T}_2$ is composed of orientation $\mathbf{R}_2$ and position $\mathbf{p}_2$. $d_m=0.1 \m$ is the optimal baseline. $\alpha_m$ is set as $5^{\circ}$.

	The neighboring keyframes selection is done by searching the optimal training image's neighbors. {Once the frame $k$ is aligned by deep CNN, the backward (or forward) keyframe seeds $k_{b}$ (or $k_{f}$) are set as $k$. The backward (or forward) direction searches the optimal (highest score) keyframe in the range $[k_{b}-\triangle k,k_{b}]$/$[k_{f},k_{f}+\triangle k]$ ($\triangle k>0$). Then the optimal keyframe is assigned as $k_{b}$ (or $k_{f}$).} This process is iteratively carried out on both directions until the joint keyframe group reaches the maximum number. \par

	\subsection{The single image locator}
	\label{Section_geometric_locator}

	The geometric locator consists of training image sparse mapping ({forward-intersection}) and testing image localization ({backward-intersection}). It localizes a single image geometrically, and both modules are simplified from BA \cite{hartley2003multiple}.\par

\begin{figure}[t]
		\centering
		\includegraphics[width=0.97\columnwidth]{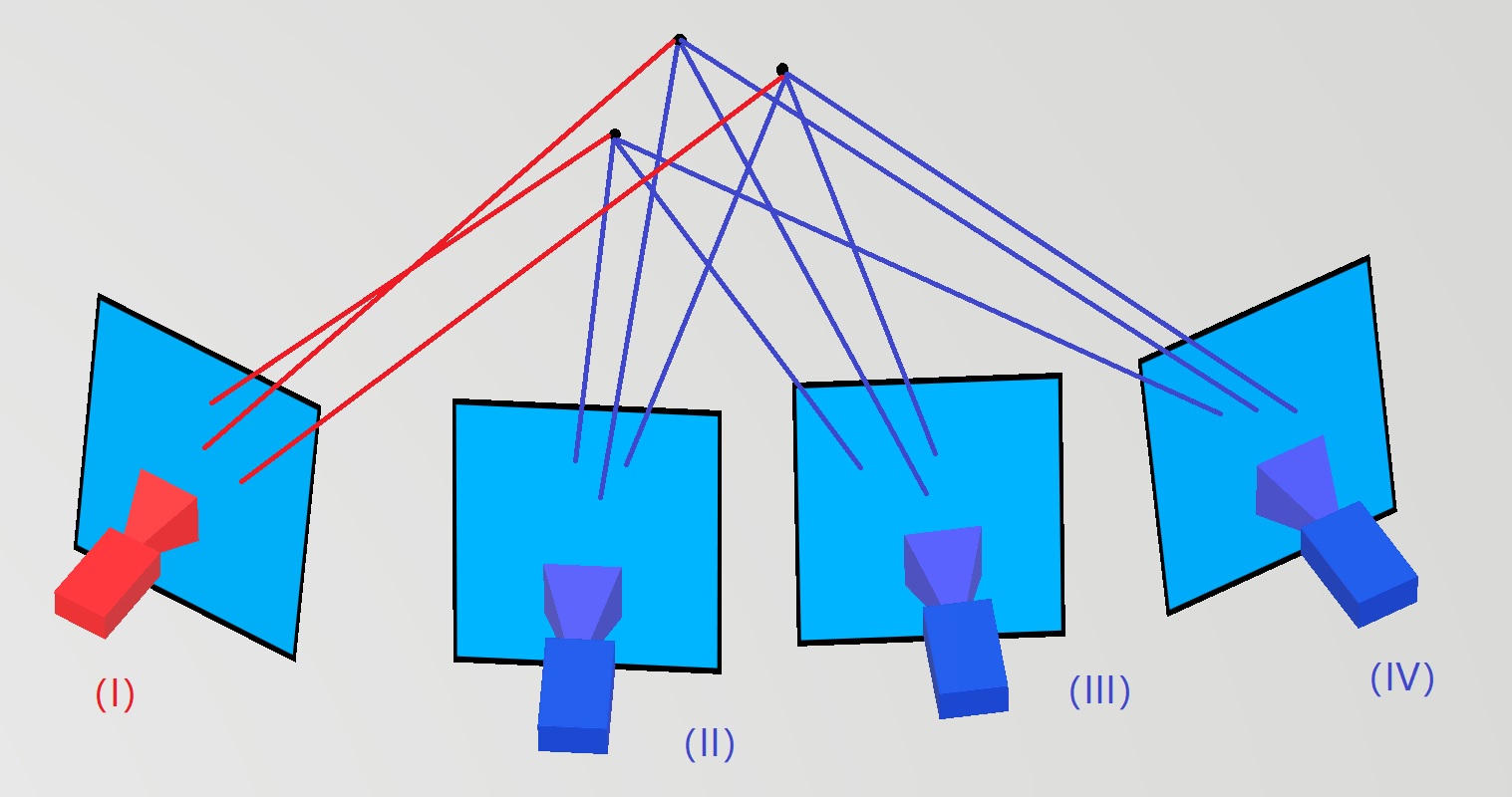}
		\caption{The figure shows the forward-intersection and the backward-intersection. The red camera (I) is the testing pose $\mathbf{T}_i$ while the three blue cameras are the training set (II, III and IV) $\mathbf{T}_{t-2}$, $\mathbf{T}_{t-1}$ and $\mathbf{T}_{t}$. Three 2D feature points are tracked on the training and testing images. The forward-intersection triangulates the global coordinates of the 3D feature points tracked in the training images using the labeled poses. After the forward-intersection, the backward-intersection estimates the pose of the testing images by minimizing the re-projection errors of the 3D triangulated feature points.}
		\label{fig_Triangulation_BA}
	\end{figure}

	Forward-intersection triangulates sparse 3D map points from the 2D tracking in the training set. The predicted most similar image and its neighbors in Section \ref{section_neighbor_select} are adopted. As Fig. \ref{fig_Triangulation_BA} shows, the predicted image, as well as its co-visible images (blue camera in Fig. \ref{fig_Triangulation_BA}), go through the SURF to be tracked with 2D key corners. All selected images are paired and matched for reliable tracking. The poses of the selected images are $\mathbf{T}_{k} \in \mathrm{SE}(3) $. Define the 2D point $j$ tracked on image $k$ as $\mathbf{x}^j_k$. Note that some points cannot be tracked on all images. Projecting the 3D point $\mathbf{X} \in \mathbb{R}^{3}$ in global coordinate to the image with pose $\mathbf{T} \in \mathrm{SE}(3)$ is\par
	
	\begin{equation*}
	(\Pi(\mathbf{T},\mathbf{X})^{\transpose} \mid 1)^{\transpose}:=\mathbf{K}\mathbf{T}^{-1}(\mathbf{X}^{\transpose} \mid 1)^{\transpose},
	\end{equation*}
	\noindent where  $\mathbf{x}:=\Pi(\mathbf{T},\mathbf{X})^{\transpose} \in \mathbb{R}^{2}$ is the 2D projection. It obtains its homogeneous version. $\mathbf{K} \in \mathbb{R}^{3\times4}$ is the camera intrinsic matrix. Forward-intersection retrieves the 3D map points $\mathbf{X}^1, \dots, \mathbf{X}^t$ by minimizing the sum re-projection residual as\par

	\begin{equation}
	\label{Eq_forward_intersection}
	\begin{split}
	\min\limits_{\mathbf{X}^1, \dots, \mathbf{X}^t} &\sum_{k\in \mathit{T}}\sum_{j\in \Omega_k} \lVert\Pi(\mathbf{T}_k,\mathbf{X}^j)-{\mathbf{x}}_k^j\lVert^2,
	\end{split}
	\end{equation}
	
	\noindent where $\mathit{T}$ is the set of all selected training images and $\Omega_k$ is the group of key points ${\mathbf{x}}_k^j$ tracked on image $k$. Huber-loss \cite{huber1992robust,kummerle2011g} is applied as the robustifier to handle outliers. All trackings are treated with equal weights. Moreover, the triangulated points with an average residual greater than threshold $\epsilon_r$ are deleted.\par

	Backward-intersection localizes the camera of the testing image $\mathbf{T}^G_{i} \in \mathrm{SE}(3)$ with the triangulated 3D points. The 3D feature points are matched with the testing image (red camera in Fig. \ref{fig_Triangulation_BA}). Define $\Omega_i$ as the group of the tracked 2D points on image $i$. $\mathbf{T}^G_i$ is estimated by minimizing the sum of re-projection residual \eqref{Eq_back_projectioin} as
	\begin{equation}
	\label{Eq_back_projectioin}
	\begin{split}
	\min\limits_{\mathbf{T}^G_{i} \in \mathrm{SE}(3)} &\sum_{j\in \Omega_{i}} \lVert\Pi(\mathbf{T}^G_{i},\mathbf{X}^j)-\tilde{\mathbf{x}}_i^{j}\lVert^2,
	\end{split}
	\end{equation}

	\noindent where $\tilde{\mathbf{x}}_i^{j}$ is the $j$th tracked points on the testing image $i$. Denote the sum of the residuals in \eqref{Eq_back_projectioin} as $\mathbf{r}_i$. Similar to forward-itersection, Huber-loss \cite{huber1992robust,kummerle2011g} is applied and all tracking are treated with equal weights (variance is set to $1$ pixel). Combining the identity variance matrix of the pixel-level and the Huber loss factors, the information matrix of each map point $\mathbf{I}(\tilde{\mathbf{x}}_i^{j})$ can be obtained. Denote $\mathcal{J}^{-1}_j(\mathbf{T}^G_{i})$ as the right Jacobian of the per-pixel residuals in~\eqref{Eq_back_projectioin} following the definition in \cite{barfoot2017state}, the information matrix of the pose $\mathbf{T}^G_{i}$ is given by
	\begin{equation*}
	\mathbf{I}(\mathbf{T}^G_{i})=\frac{1}{\sigma(\overline{\mathbf{r}}_i)^2} \sum_{j\in \Omega_{i}}\mathcal{J}^{-1}_j(\mathbf{T}^G_{i})^\transpose \mathbf{I}(\tilde{\mathbf{x}}_i^{j}) \mathcal{J}^{-1}_j(\mathbf{T}^G_{i}).
	\end{equation*}
	We follow~\cite{polic2017camera} to approximate the average residual variance $\sigma(\overline{\mathbf{r}}_i)^2$ as 

	\begin{equation*}
	\sigma(\overline{\mathbf{r}}_i)^2 \approx\lVert\mathbf{r}_i\lVert^{2} /(2 \mathrm{p}-1),
	\end{equation*}

	\noindent where $\mathrm{p}$ is the number of tracking in $\Omega_{i}$. With the information matrix $\mathbf{I}(\mathbf{T}^G_{i})$, the covariance matrix ${\mathbf{C}}^G_{i} \in \mathbb{R}^{6\times6}$ can be obtained as
	\begin{equation}
	\label{Eq_var_sfm}
	{\mathbf{C}}^G_{i}=\mathbf{I}(\mathbf{T}^G_{i})^{-1}.
	\end{equation}

	\subsection{Probabilistic motion model}
	As \citet{patel2018contextualnet} pointed out, single image-based localization suffers from predictions that deviate significantly from consecutive predictions. Sequential predictions can alleviate this issue as it uses information from multiple images. Researchers developed different techniques that exploits sequential information using both deep CNN based methods~\cite{patel2018contextualnet,valada2018deep,radwan2018vlocnet++,tian20203d} and geometry perspective~\cite{brahmbhatt2018geometry, yang2020d3vo}. We applied short-term robot motion model~\cite{thrun2002probabilistic,barfoot2017state} to constrain the consecutive predictions. The short-term robot motion model is often categorized as Brownian, constant velocity, constant acceleration, and constant turn \cite{thrun2002probabilistic,shi2002speed}. We employ a constant velocity-based probabilistic motion model to fit the short-term robot motion. The motion model is first fitted with the short-term historical prediction and then applied to predict the testing image's pose and the uncertainty. Following the constant motion interpolation (in $\mathrm{SE}(3)$) in~\cite{barfoot2017state}, we define the short-term constant motion $\delta_ \mathbf{T} \in \mathrm{SE}(3)$ and optimized the objective function as 
	\begin{equation}
	\label{Eq_motionmodel_position}
	\begin{split}
	\min\limits_{\delta_ \mathbf{T},\mathbf{T}^{M}_i} &\sum_{j=1}^{t} 
	\lVert \operatorname{d}_{\mathfrak{se}(3)}({\mathbf{T}^F_{i-j}},\mathbf{T}^{M}_i\delta_ {\mathbf{T}}^{j-1}) \lVert^2,
	\end{split}
	\end{equation}
	
	\noindent where $t$ is the number of historical images in the window of the motion model, $\mathbf{T}^{M}_i \in \mathrm{SE(3)}$ is the $i$th estimated pose. The basic idea is to search for the optimal $\mathbf{T}^{M}_i$ and constant motion $\delta_ \mathbf{T}$ to fit the consecutive predictions. After the estimation of the probabilistic motion model \eqref{Eq_motionmodel_position}, the one-step prediction of robot pose can be achieved.\par

	The covariance matrix (diagonal) of $\mathbf{T}^M_i$ from the probabilistic motion model, which is ${\mathbf{C}}^M_{i}$, is calculated statistically. It describes the fitting accuracy of the probabilistic motion model. The uncertainty only quantifies the short-term epistemic uncertainty and suffers from uncertainty drift in the long run. Precisely, if the uncertainties of the motion model are propagated, they tend to be over-confident.\par

	\subsection{Fusion of the geometric locator and the motion model}

	The predictions of the geometric locator and the probabilistic motion model are fused with the quantified uncertainties following \cite{barfoot2014associating} as 

	\begin{equation}
	\label{Eq_fusion}
	\begin{split}
	\min\limits_{\mathbf{T}^{F}_i} 
	&\operatorname{d}_{\mathfrak{se}(3)}(\mathbf{T}_i^M,\mathbf{T}^{F}_i)^\transpose \left({\mathbf{C}}^M_{i}\right)^{-1} \operatorname{d}_{\mathfrak{se}(3)}(\mathbf{T}_i^M,\mathbf{T}^{F}_i)\\
	&+\operatorname{d}_{\mathfrak{se}(3)}(\mathbf{T}_i^G,\mathbf{T}^{F}_i)^\transpose \left({\mathbf{C}}^G_{i}\right)^{-1} \operatorname{d}_{\mathfrak{se}(3)}(\mathbf{T}_i^G,\mathbf{T}^{F}_i),
	\end{split}
	\end{equation}

	\begin{equation*}
	\begin{aligned}  
	{\mathbf{C}}^F_{i}\!=\!
	\left({\mathcal{J}_{r}^{M}}^{-\transpose} \left({\mathbf{C}}^M_{i}\right)^{-1}
	{{\mathcal{J}_{r}^{}}^{M}}^{-1}+{\mathcal{J}_{r}^{G}}^{-\transpose}\! \left({\mathbf{C}}^G_{i}\right)^{-1} {\mathcal{J}_{r}^{G}}^{^{-1}}\!\right)^{-1}\!,
	\end{aligned}  
	\end{equation*}

	\noindent where ${\mathcal{J}^M_{r}}^{-1}$ and ${\mathcal{J}^G_{r}}^{-1}$ are the right Jacobians of objective function \eqref{Eq_fusion} regarding $\mathbf{T}^M_{i}$ and $\mathbf{T}^G_{i}$ following the definition in \cite{barfoot2017state}. To filter outliers from the single image locator, the isometric standard deviation $3\sigma({\mathbf{T}_i^G})|_p$ and $3\sigma({\mathbf{T}_i^G})|_r$ are enforced for filtering $\mathbf{T}_i^G$. If the positional or orientational predictions of $\mathbf{T}_i^G$ exceeds the bounds, $\mathbf{T}_i^G$ is identified as failure and the motion model pose prediction $\mathbf{T}^M_{i}$ is assigned to $\mathbf{T}^{F}_i$. The isometric standard deviation ${\sigma}(\mathbf{T}^G_{i})|_p$ and ${\sigma}(\mathbf{T}^G_{i})|_r$ are

	\begin{equation*}
	\label{Eq_uncertainty_p_iso}
	{\sigma}(\mathbf{T}^G_{i})|_p = \frac{1}{3}\sum_{i=1}^3 \sqrt{{\mathbf{C}}^G_{i}|_p}, \  
	{\sigma}(\mathbf{T}^G_{i})|_r = \Theta(\exp(\sqrt{{\mathbf{C}}^G_{i}|_r})^\wedge),
	\end{equation*}

	\noindent where scalar ${\sigma}(\mathbf{T}^G_{i})|_p$ represents position's ellipsoid uncertainty. Similarly, scalar ${\sigma}(\mathbf{T}^G_{i})|_r$ is the orientational uncertainty.

	The motion model provides an independent and cheap way to rectify the outliers. Unlike the global positioning of the geometric locator, the probabilistic motion model retrieves the local pose with the estimated short-term historical poses. Its uncertainty is only valid in the short term. Assuming the error of the pose from the geometric locator follows the Gaussian distribution if the motion model prediction deviates the geometric locator for more than 3-$\sigma$ bound, the prediction of the geometric locator is discarded. The large difference indicates unreasonable abrupt pose prediction. Other filtering methods, such as the extended Kalman filter, are unsuitable because the geometric locator yields consecutive outliers.

	\section{Experiments}
   We evaluated our method on three data sets (one synthetic and two real-world). The synthetic data set was generated in a virtual warehouse using Isaac Sim~\cite{issacsdk} engine. The real-world public data set used for our experiments were 7Scenes~\cite{shotton2013scene} and TUM RGB-D data sets~\cite{sturm12iros}. We also used a Monte-Carlo experiment to validate the uncertainty quantification proposed in~\eqref{Eq_var_sfm}.

	\subsection{Monte-Carlo synthetic experiment}

	To validate the uncertainty quantification in \eqref{Eq_var_sfm}, a Monte-Carlo experiment was conducted to test the geometric locator. It follows the scenario in Fig. \ref{fig_Triangulation_BA}. $119$ 3D points were partially projected on the images of the 4 camera poses and the target camera. Gaussian noises (i.i.d.) with variance $\sigma_p^2$ were exerted on all 2D tracking. The coverage rate is defined as the difference between the estimated pose and the ground truth pose $\lVert\mathbf{T}^G_{k}|_p-\overline{\mathbf{T}}^G|_p\rVert_2$ and $\Theta(\mathbf{T}^G_{k}|_r^\transpose\overline{\mathbf{T}}^G|_r)$ within $[-1.96
	{\sigma}(\mathbf{T}^G_{i})|_p,1.96
	{\sigma}(\mathbf{T}^G_{i})|_p]$ and $[-1.96{\sigma}(\mathbf{T}^G_{i})|_r,1.96{\sigma}(\mathbf{T}^G_{i})|_r]$, where $k$ is the $k$th Monte-Carlo experiment and $\overline{\mathbf{T}}^G$ is the ground truth. To further evaluate the efficiency with the presence of noise in pose labeling, we corrupt pose labels using simulated noise to replicate a more realistic scenario. Table \ref{Table_Monte_Carlo} shows the coverage rate using different $\sigma_p$ and different pose label noises. The ideal coverage rate is $95\%$. \par  

	\begin{table}[]
		\centering
		\scriptsize
		\caption{The table shows the Monte-Carlo experiments with 1000 trials. We show the coverage rate of different levels of tracking noises. We also tested adding isometric pose noises to pose labels (cells are standard deviation of the noises). }
		\resizebox{\columnwidth}{!}{
			\begin{tabular}{m{1.6cm}<{\centering}|m{0.85cm}<{\centering}|m{0.85cm}<{\centering}|m{0.85cm}<{\centering}|m{0.85cm}<{\centering}|m{0.85cm}<{\centering} |m{0.85cm}<{\centering}}
				%\begin{tabular}{|c|c|c|c|c|c|c|}
				% 			\hline
				\begin{tabular}[c]{@{}c@{}}Pose noise\end{tabular} & \multicolumn{2}{c|}{$0\m$, $0^{\circ}$} & \multicolumn{2}{c|}{$0.01\m$, $0.5^{\circ}$} & \multicolumn{2}{c}{$0.03\m$, $1^{\circ}$} \\ \midrule
				\begin{tabular}[c]{@{}c@{}}$\sigma_p$ (pixel)\end{tabular}   & 1               & 2             & 1                 & 2                & 1                & 2               \\ \midrule
				Angular       & $95.2\%$          & $94.6\%$         & $84.6\%$             & $79.8\%$            & $49.8\%$            & $62.4\%$           \\
				Positional    & $96.0\%$           & $94.2\%$         & $92.8\%$             & $91.0\%$            & $74.4\%$            & $80.6\%$           \\ \bottomrule
			\end{tabular}
		}
		\label{Table_Monte_Carlo}
	\end{table}

	\subsection{Data set}

\begin{figure}[]
		\centering
		\includegraphics[width=0.47\textwidth]{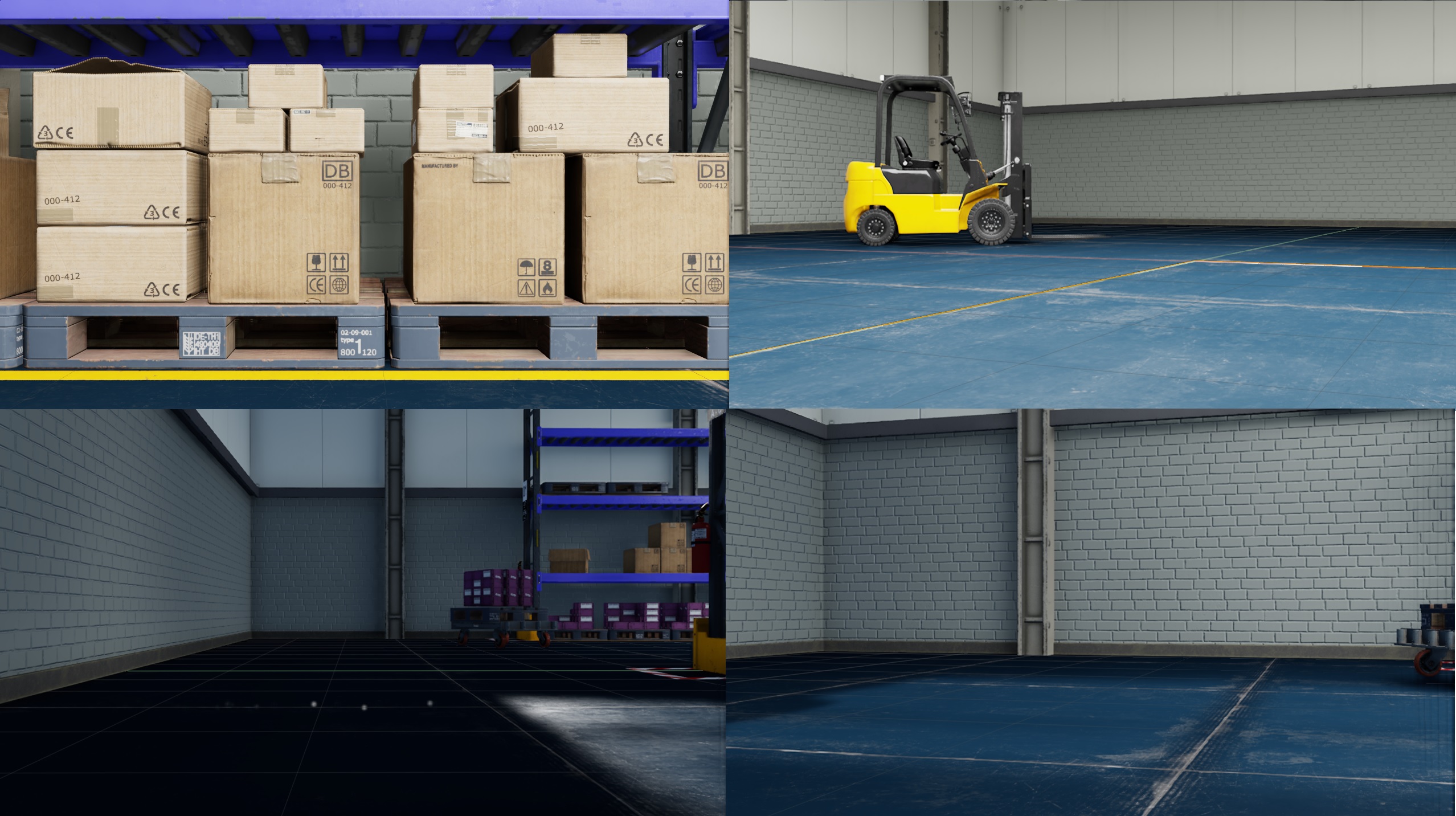}
		\caption{Shown are the examples of the synthetic data set generated using Isaac Sim simulator. The upper two and lower two figures are taken in diffuse lighting and indoor lighting respectively.}
		\label{fig_sample_isaac}
		\vspace{-3mm}
	\end{figure}

	Our method was tested on the synthetic data set generated using Isaac Sim engine~\cite{issacsdk}, that is photorealistic and physically accurate. The monocular images captured by the camera mounted on the robot were logged along with the corresponding poses while the robot maneuvered in the warehouse. Two data sets were generated in two different virtual environments with different levels of complexity. The synthesized images were in $1280\times 720$. Fig. \ref{fig_sample_isaac} shows sample images.\par

	The proposed approach was also tested on the small-range 7Scenes \cite{shotton2013scene} indoor data set, which was collected with a handheld Kinect RGB-D camera at the resolution of $480\times 640$. The pose labels were generated by implementing the KinectFusion \cite{newcombe2011kinectfusion} SLAM on the RGB-D images. The monocular images contain motion blur, flat surfaces, and various lighting conditions. The TUM RGB-D data set \cite{sturm12iros}, which was designed to test the performance of RGB-D SLAM algorithms, was also adopted. In TUM RGB-D data set, a motion capture system was used to track the reflectance markers on the Kinect camera accurately, and the intrinsic parameter was calibrated. Therefore, the accuracy of the pose label and intrinsic parameters are substantially improved compared to the 7Scenes. The image resolution is $480\times 640$ and contains apparent challenges including motion blur, flat surfaces, and various lighting conditions. We chose the last 100 images as the testing data set and the remaining as the training data set. Data sets ``freiburg1 desk2'', ``freiburg1 floor'', ``freiburg1 room'', ``freiburg2 360 hemisphere'', ``freiburg2 360 kidnap'', ``freiburg2 large with loop'' were chosen since most of the last 100 frames were revisited in the training set. We only used RGB images and the pose labels. \par

	\begin{figure}[]
		\centering
		\includegraphics[width=1.\columnwidth]{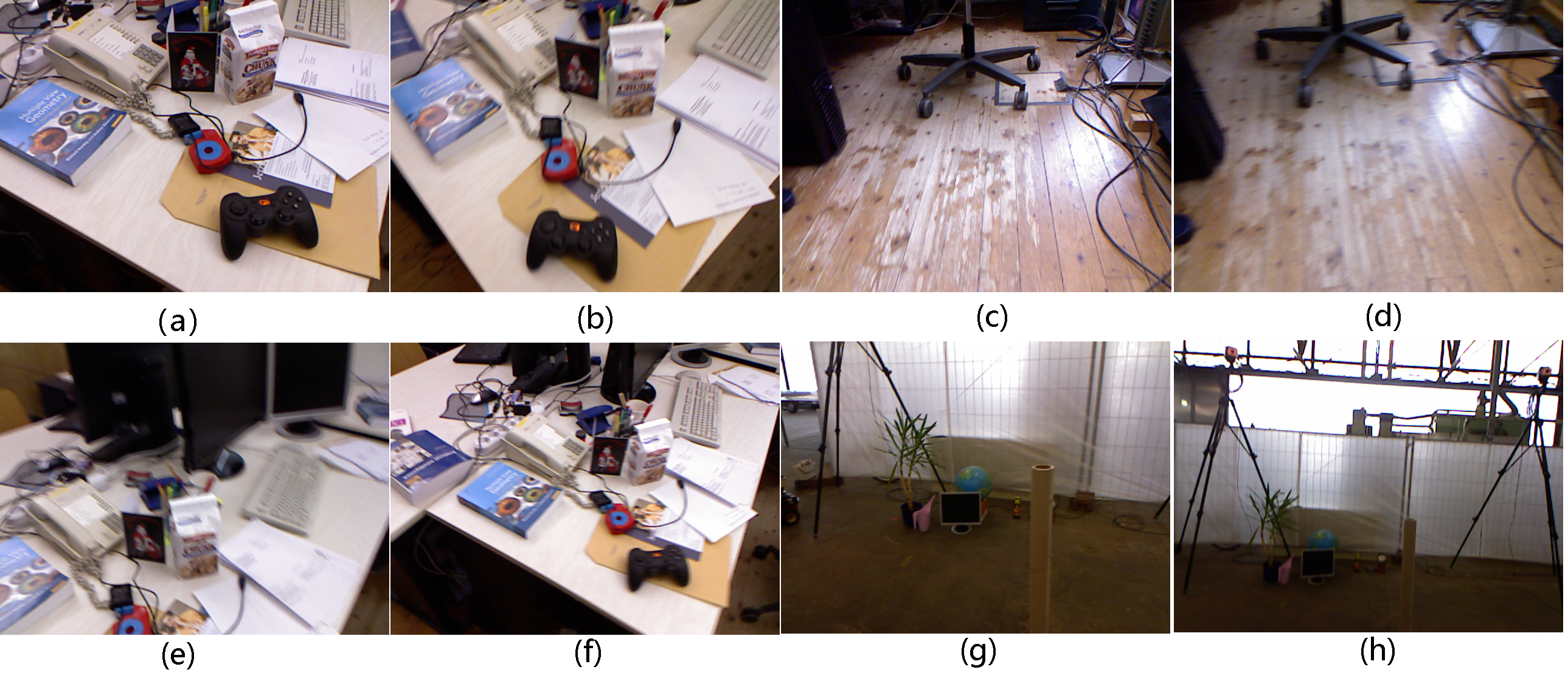}
		\caption{(a,c,e,g) are the sample testing images in the TUM data sets ``desk2'', ``floor'', ``room'' and ``large\_with\_loop''. (b,d,f,h) are the corresponding most visually similar images in the training data set predicted by the image descriptor network. We achieve the accuracy $(0.02\m, 1.51^\circ)$, $(0.01\m, 2.45^\circ)$, $(0.03\m,1.37^\circ)$ and $(0.55\m, 6.54^\circ)$ in contrast to the predictions $(0.33\m, 5.53^\circ)$, $(0.05\m, 1.66^\circ)$, $(0.35\m, 2.27^\circ)$ and $(0.97\m, 7.99^\circ)$ from PoseNet.}
		\label{fig_viewing_angle}
	\end{figure}

	\begin{table}[]
		\scriptsize
		\caption{The table shows the median localization errors on the synthetic data set generated using Isaac Sim. ``Warehouse1'' is the normal illumination, and ``warehouse2'' is in dark illumination. ``BoW+Locator'' is the integration of BoW with our geometric Locator.}
		\setlength\tabcolsep{3.5pt}
		\begin{tabular}{l|p{0.91cm}<{\centering}|p{0.91cm}<{\centering}|p{1.31cm}<{\centering}|p{1.37cm}<{\centering}|p{1.45cm}<{\centering}}
			& \begin{tabular}[c]{@{}c@{}}BoW +\\ Locator\end{tabular} & PoseNet   & ContexualNet & \begin{tabular}[c]{@{}c@{}}Our method\\ (Single image)\end{tabular}  & \begin{tabular}[c]{@{}c@{}}Our method\\ (Multiple image)\end{tabular}\\
			\midrule
			\begin{tabular}[c]{@{}l@{}} Warehouse1 \\  (simple) \end{tabular} & \begin{tabular}[c]{@{}l@{}}$0.55\m$,\\  $3.09^{\circ}$\end{tabular}& \begin{tabular}[c]{@{}l@{}}$0.24\m$,\\  $2.26^{\circ}$\end{tabular} &  \begin{tabular}[c]{@{}l@{}}$0.29\m$,\\  $3.65^{\circ}$\end{tabular}  & \begin{tabular}[c]{@{}l@{}}$0.21\m$,\\  \bm{$1.95^{\circ}$}\end{tabular}  & 
			\begin{tabular}[c]{@{}l@{}}\bm{$0.19\m$},\\  $2.01^{\circ}$\end{tabular}         \\
			\begin{tabular}[c]{@{}l@{}} Warehouse2 \\  (complex) \end{tabular} &  \begin{tabular}[c]{@{}l@{}}$3.12\m$,\\  $13.26^{\circ}$\end{tabular} &  \begin{tabular}[c]{@{}l@{}}$3.58\m$,\\  $21.77^{\circ}$\end{tabular}  &    
			\begin{tabular}[c]{@{}l@{}}$1.89\m$,\\  $14.32^{\circ}$\end{tabular}
			& \begin{tabular}[c]{@{}l@{}}$0.79\m$,\\  $10.42^{\circ}$\end{tabular}
			& \begin{tabular}[c]{@{}l@{}}\bm{$0.51\m$},\\  \bm{$8.24^{\circ}$}\end{tabular}   \\
			\bottomrule
		\end{tabular}
		\label{Table_Isaac_sim}
	\end{table}

	\begin{table*}[t]
		\centering
		\scriptsize
		\caption{The table shows the median localization errors of SCORE forest \cite{shotton2013scene}, directional PoseNet \cite{kendall2015posenet}, Bayesian PoseNet \cite{kendall2016modelling}, the ContexturalNet \cite{patel2018contextualnet} and the 3D geometry-aware network \cite{tian20203d} in the 7Scenes data set.}
		\setlength\tabcolsep{5pt}
		\begin{tabular}{l|p{1.6cm}<{\raggedleft}|p{1.75cm}<{\centering}|p{1.75cm}<{\centering}|p{1.75cm}<{\centering}|p{1.75cm}<{\centering}|p{1.75cm}<{\centering}|p{1.75cm}<{\centering}|p{1.75cm}<{\centering} p{2.2cm}<{\centering}|}
			& \begin{tabular}[c]{@{}c@{}}Score Forest\\ (RGB-D)\end{tabular} & PoseNet    & \begin{tabular}[c]{@{}c@{}}Directional\\ PoseNet\end{tabular}  & \begin{tabular}[c]{@{}c@{}}Bayesian\\ PoseNet\end{tabular} & ContexualNet & \begin{tabular}[c]{@{}c@{}}Geometry-\\ aware DNN~\cite{tian20203d}\end{tabular}  & \begin{tabular}[c]{@{}c@{}}Our method\\ (Single image)\end{tabular} & \begin{tabular}[c]{@{}c@{}}Our method\\ (Multiple model)\end{tabular} \\ \midrule 
			Chess       & $0.03\m$, $0.66^{\circ}$                                        & $0.32\m$, $8.12^{\circ}$  & $0.24\m$, $5.77^{\circ}$           & $0.37\m$, $7.24^{\circ}$       & $0.15\m$, $6.12^{\circ}$    & $\bm{0.09\m}$, $4.39^{\circ}$    & $0.12\m$, $4.06^{\circ}$                                                        & $0.13\m$, $\bm{3.87^{\circ}}$                                                       \\
			Fire        & $0.05\m$, $1.50^{\circ}$                                        & $0.47\m$, $14.40^{\circ}$  & $0.34\m$, $11.90^{\circ}$          & $0.43\m$, $13.70^{\circ}$       & $0.16\m$, $10.93^{\circ}$ & $0.25\m$, $4.79^{\circ}$   & $\bm{0.08\m}$, $4.11^{\circ}$                                                          & $\bm{0.08\m}$, $\bm{4.04^{\circ}}$                                                        \\
			Heads       & $0.06\m$, $5.50^{\circ}$                                        & $0.29\m$, $12.00^{\circ}$ & $0.21\m$, $13.70^{\circ}$          & $0.29\m$, $12.00^{\circ}$        & $0.25\m$, $13.20^{\circ}$ & $0.14\m$, $12.56^{\circ}$   & $0.08\m$, $\bm{4.42^{\circ}}$                                                        & $\bm{0.05\m}$, $4.50^{\circ}$                                                       \\
			Office      & $0.04\m$, $0.78^{\circ}$                                        & $0.48\m$, $7,68^{\circ}$  & $0.30\m$, $8.08^{\circ}$           & $0.48\m$, $7.68^{\circ}$        & $0.25\m$, $7.45^{\circ}$  & $0.17\m$, $\bm{6.46^{\circ}}$   & $0.23\m$, $9.64^{\circ}$                                                      & $\bm{0.14\m}$, $7.31^{\circ}$                                                       \\
			Pumpkin     & $0.04\m$, $0.68^{\circ}$                                        & $0.47\m$, $8.42^{\circ}$  & $0.33\m$, $7.00^{\circ}$           & $0.47\m$, $8.42^{\circ}$        & $0.26\m$, $6.62^{\circ}$  & $0.19\m$, $\bm{5.91^{\circ}}$   & $0.25\m$, $8.79^{\circ}$                                                       & $\bm{0.16\m}$, $6.30^{\circ}$                                                       \\
			\begin{tabular}[c]{@{}c@{}}Red Kitchen\end{tabular} & $0.04\m$, $0.76^{\circ}$                                        & $0.59\m$, $8.43^{\circ}$  & $0.37\m$, $8.83^{\circ}$           & $0.59\m$, $8.64^{\circ}$        & $0.20\m$, $6.97^{\circ}$  & $0.21\m$, $6.47^{\circ}$    & $\bm{0.02\m}$, $\bm{1.10^{\circ}}$                                                        & $\bm{0.02\m}$, $1.40^{\circ}$                                                       \\
			Stairs      & $0.03\m$, $1.32^{\circ}$                                        & $0.47\m$, $13.80^{\circ}$ & $0.40\m$, $13.70^{\circ}$          & $0.47\m$, $13.80^{\circ}$        & $0.17\m$, $10.83^{\circ}$ & $0.26\m$, $11.51^{\circ}$  & $0.11\m$, $\bm{5.22^{\circ}}$                                                       & $\bm{0.08\m}$, $5.40^{\circ}$                                                       \\ \bottomrule
		\end{tabular}
		% 		}
		\label{Table_quantitative_compare}
	\end{table*}

	\begin{table}[]
		\centering
		\scriptsize
		\caption{The table shows the median localization errors on the TUM data~set.}
		\begin{tabular}{l|p{0.93cm}<{\centering}|p{1.21cm}<{\centering}|p{1.31cm}<{\centering}|p{1.65cm}<{\centering}}
			& PoseNet        & ContexualNet & \begin{tabular}[c]{@{}c@{}}Our method\\ (Single image)\end{tabular}  & \begin{tabular}[c]{@{}c@{}}Our method\\ (Multiple image)\end{tabular}  \\ \midrule 
			desk2                                      & \begin{tabular}[c]{@{}l@{}}$0.21\m$,\\  $15.81^{\circ}$\end{tabular} & \begin{tabular}[c]{@{}l@{}}$0.05\m$, \\ $10.75^{\circ}$\end{tabular}          & \begin{tabular}[c]{@{}l@{}}$0.07\m$, \\ $\bm{2.91^{\circ}}$\end{tabular}                                                                    & \begin{tabular}[c]{@{}l@{}}$\bm{0.03\m}$, \\ $2.74^{\circ}$\end{tabular}                                                                      \\
			floor                                     & \begin{tabular}[c]{@{}l@{}}$0.16\m$, \\ $1.87^{\circ}$\end{tabular}  & \begin{tabular}[c]{@{}l@{}}$0.05\m$, \\ $2.86^{\circ}$\end{tabular}           & \begin{tabular}[c]{@{}l@{}}$\bm{0.02\m}$, \\ $0.96^{\circ}$\end{tabular}                                                                    & \begin{tabular}[c]{@{}l@{}}$\bm{0.02\m}$, \\ $\bm{0.84^{\circ}}$\end{tabular}                                                                 \\
			room                                      & \begin{tabular}[c]{@{}l@{}}$0.19\m$, \\ $7.56^{\circ}$\end{tabular}  & \begin{tabular}[c]{@{}l@{}}$0.06\m$, \\ $4.99^{\circ}$\end{tabular}           & \begin{tabular}[c]{@{}l@{}}$\bm{0.02\m}$, \\ $1.44^{\circ}$\end{tabular}                                                                    & \begin{tabular}[c]{@{}l@{}}$0.03\m$, \\ $\bm{1.21^{\circ}}$\end{tabular}                                                                      \\
			360\_hemisphere                           & \begin{tabular}[c]{@{}l@{}}$1.00\m$, \\ $8.53^{\circ}$\end{tabular}  & \begin{tabular}[c]{@{}l@{}}$\bm{0.62\m}$, \\ $\bm{7.69^{\circ}}$\end{tabular} & \begin{tabular}[c]{@{}l@{}}$1.35\m$, \\ $8.48^{\circ}$\end{tabular}                                                                         & \begin{tabular}[c]{@{}l@{}}$1.14\m$, \\ $8.41^{\circ}$\end{tabular}                                                                           \\
			large\_with\_loop                        & \begin{tabular}[c]{@{}l@{}}$0.58\m$, \\ $9.11^{\circ}$\end{tabular}  & \begin{tabular}[c]{@{}l@{}}$0.41\m$, \\ $5.07^{\circ}$\end{tabular}           & \begin{tabular}[c]{@{}l@{}}$0.38\m$, \\ $\bm{3.55^{\circ}}$\end{tabular}                                                                    & \begin{tabular}[c]{@{}l@{}}$\bm{0.27\m}$, \\ $3.62^{\circ}$\end{tabular}            \\ \bottomrule
		\end{tabular}
		\label{Table_TUM_dataset}
	\end{table}

\begin{table}[]
    \centering
	\scriptsize
	\caption{The table compares our method (single image) and PoseNet on the downsampled synthetic warehouse (different ratio of downsampling) data set.  Values are in median errors.}
\begin{tabular}{l|p{0.64cm}<{\centering}|p{0.64cm}<{\centering}|p{0.64cm}<{\centering}|p{0.64cm}<{\centering}|p{0.64cm}<{\centering}|p{0.64cm}<{\centering}}
\multicolumn{1}{c|}{} & 100\% & 75\% & 50\% & 33\% & 25\% & 12\% \\ \midrule
Ours (Simple)    & 0.21m  & 0.21m & 0.22m & 0.24m & 0.23m & 0.25m    \\
Ours (Complex)   & 0.79m  & 0.79m & 0.83m & 0.88m & 0.97m & 0.94m    \\
PoseNet (Simple)    & 0.24m & 0.25m & 0.30m & 0.31m & 0.34m & 0.39m    \\
PoseNet (Complex)   & 3.58m & 3.54m & 4.66m & 4.32m & 4.73m & 4.91m    \\ \bottomrule
\end{tabular}
\label{Table_train_data}
\end{table}

	\subsection{Experiment setting}

	Following settings were applied in the training process. In each epoch, 128 positive image pairs and the same amount of negative pairs were randomly selected. The model was trained for $500$ epochs with the learning rate of $0.001$ and the batch size of $128$. The training was achieved by minimizing the contrastive loss; the weights were optimized using Adaptive movement estimation (Adam)~\cite{kingma:2014}. Margin $\mathrm{m}$ in the contrastive loss function is set as $1$.\ Tensorflow libraries~\cite{abadi2016tensorflow} were employed for training, and the model was trained on NVIDIA Titan X GPUs. The model was initialized using the weights trained on the Places~\cite{zhou:2014} data set.\par 
	
	In the keyframe selection, the searching range $\triangle k$ is set to 100. Standard deviation $\ell=0.2$. $\alpha_{m}$ is set to 3 degrees. Seven keyframes are selected (3 backward and 3 forward).\par
	
	In the geometric locator, the threshold of the re-projection error $\eta_r$ was set to $5$. $t=4$ historical images were selected for the motion model. The proposed method was compared with the PoseNet \cite{kendall2015posenet}, ContextualNet \cite{patel2018contextualnet}, Bayesian PoseNet \cite{kendall2016modelling}, the geometry-aware DNN~\cite{tian20203d} on the 7Scenes data. It should be noted that the results for PoseNet, ContextualNet, Bayesian PoseNet, and the geometry-aware DNN~\cite{tian20203d} on the 7Scenes data set were directly cited from the published papers. For TUM and Isaac Sim data sets, we compared the results with the PoseNet \cite{kendall2015posenet} and ContextualNet \cite{patel2018contextualnet}. The accuracy comparisons are presented in Tables \ref{Table_Isaac_sim}, \ref{Table_quantitative_compare} and \ref{Table_TUM_dataset}. Each image is processed individually (with several neighboring predictions) and sequential processing is not required.\par
	
	We note that monocular SLAM cannot be implemented on the 3 data sets as the initialization fails due to the predominant rotational motion. To compare with the prior-free BoW (the backbone of the relocalization module in ORB-SLAM series), we implemented BoW for image alignment and adopted our proposed geometric locator for 6 DoF pose estimation, results of which are presented in Table~\ref{Table_Isaac_sim}. Moreover, in the first few starting frames, only single image locator is used.\par

	\section{Discussion and Limitations}

	\subsection{Performance of the proposed method}
	
	The results of Monte-Carlo experiment listed in Table \ref{Table_Monte_Carlo} shows \eqref{Eq_var_sfm} is accurate when pose labels are noise-free. The accuracy of the estimated uncertainty deteriorates as the pose label noise increases. We could not evaluate the probabilistic motion model using the Monte-Carlo tests as the accuracy is dependent on the validity of the constant displacement and turning assumption. \par

	Table \ref{Table_Isaac_sim}, \ref{Table_quantitative_compare} and \ref{Table_TUM_dataset} demonstrate our proposed method significantly outperforms the baseline methods on the synthetic, 7Scenes and TUM data sets. It can partially be attributed to the abundant texture in all three data sets. Adequate accurate and robust corner points contribute to the good performance of the method on all data sets. More importantly, the inferior performance of the image-to-pose methods is due to the different viewing angles between the training and testing image sets. Fig. \ref{fig_viewing_angle} reveals several sample testing images and their visually similar correspondences in the training set. The figures indicate that the scenarios were observed from different viewing angles. Our proposed method substantially outperforms image-to-pose fitting methods in these cases due to the efficiency of the BA and the dissimilarity between the training and testing images. Learning-based methods cannot achieve similar performance in this situation. The BA localization, however, obtains the average residual less than 1 pixel after the optimization. \par

	The motion model further addresses the misclassification issues of the Siamese network, which are caused by blurry images and textureless regions. The uncertainty estimated by the motion model, along with the geometric locator, assists in filtering incorrect results and improving the overall results.\par

	Further, our proposed method is more suitable to be applied in the scenario where there are fewer training data and abundant textures. Conventional learning-based methods heavily rely on the size of the training data. We tested PoseNet and our method with downsampled training data set as listed in Table~\ref{Table_train_data}, which shows the feasibility of our approach with smaller data set. Moreover, as Table~\ref{Table_TUM_dataset} indicates, the deep CNN-based model is not well trained with limited data (540 and 1261 training samples for ``desk2'' and ``room'' data set respectively). Take the training samples 540, e.g., the Siamese training procedure can provide $6480$ samples at most (positive and negative pairs). Furthermore, our method's image similarity network task is easier to achieve than the image-to-pose fitting.\par

	\subsection{Accuracy of the image descriptor network}

	``Single image'' results in Table \ref{Table_Isaac_sim}, \ref{Table_quantitative_compare} and \ref{Table_TUM_dataset} validate the efficiency of the image descriptor network. The accuracy of the geometric locator heavily relies on testing and training image similarity alignment. The geometric locator fails if the overlap between the two images is sufficiently small that there are not enough valid features to be tracked. We measure the median distance between the testing and aligned training image in TUM data set, which are: ($0.90\m$, $39.19^{\circ}$), ($0.72\m$, $20.41^{\circ}$), ($2.06\m$, $37.76^{\circ}$), ($1.17\m$, $58.76^{\circ}$), ($0.93\m$, $76.01^{\circ}$) and ($2.15\m$, $58.71^{\circ}$) respectively.

	\subsection{The benefit of introducing the geometric uncertainty}

	One critical drawback of the deep image-to-pose fitting is that it relies on the domain coverage between the training and testing data. The Bayesian network \cite{kendall2016modelling} estimates the uncertainty with its dropout approximated framework and passively ``provides a measure of similarity between a testing image and the data set’s training image''. Our method actively searches for key features to align the images when the testing image and training image are observed from different viewing directions. Moreover, the aleatoric uncertainty from the re-projection errors geometrically depicts the confidence of the estimated pose well. The uncertainty addresses the quality of the testing image and shows geometric confidence. In the experiments, images with blurriness are assigned with high uncertainty, while images with different viewing directions maintain high confidence. The wrongly classified images can also be identified if the prediction deviates from the motion model prediction significantly.\par

	With the geometric uncertainty, the fusion module enhances the robustness of the localization. The motion model provides complimentary short-term inertial information when the geometric method does not perform well (indicated by the uncertainty). As demonstrated in the three experiments, the motion model cannot improve the performance when the geometric localization works well. However, when the geometric locator performs poorly (``Office'' and ``Pumpkin'' (7Scenes), ``Warehouse2'' (simulation), e.g.), the motion model substantially improves the accuracy. \par 

	\subsection{Limitations}
	The proposed method has two drawbacks. First, the geometric locator suffers from illumination changes. The matched corner points are unreliable if the training image and the aligned testing image have very different illumination. We observed satisfactory performance in the indoor scenarios with relatively uniform illumination and poor performance in outdoor environments because of heavy influence by sunlight. Moreover, this approach requires the known camera intrinsic parameter. The two drawbacks indicate our proposed method and the image-to-pose learning-based methods \cite{kendall2015posenet,patel2018contextualnet} are complementary. Although both our proposed method and the work of~\citet{tian20203d} address the geometric constraint, experiments show that our method is more efficient given the correct camera parameter. It remains unclear which method is more suitable when the camera intrinsic parameter is unknown, and this problem is an interesting future work direction. \par

	\section{Conclusions}
    
    We developed a geometry-aware image-based localization system for indoor environments using an image descriptor network, a geometric locator, a motion model, and an uncertainty-based pose fusion. The proposed framework has higher accuracy and is even well suited for scenarios where the data sets are small. We extensively tested our method using synthetic and publicly available data sets and provided comparisons with other learning and feature-based methods.

	In the future, we plan to extend the algorithm to localize any image with unknown intrinsic parameters. We will also explore developing a fully automatic localization failure detection mechanism.\par

	%%%%%%%%%%%%%%%%%%%%%%%%%%%
% 	\vspace{-4mm}
	\section*{Acknowledgment}
	Toyota Research Institute provided funds to support this work. Funding for M. Ghaffari was in part provided by NSF Award No. 2118818. We would also like to acknowledge FX Palo Alto Laboratory Inc. where this work initially started.
% 	\vspace{-3mm}
	\balance
	{\small
		\bibliographystyle{IEEEtranN}
		\bibliography{bib/strings-abrv,bib/ieee-abrv,reference}

% Generated by IEEEtranN.bst, version: 1.14 (2015/08/26)
\begin{thebibliography}{43}
\providecommand{\natexlab}[1]{#1}
\providecommand{\url}[1]{#1}
\csname url@samestyle\endcsname
\providecommand{\newblock}{\relax}
\providecommand{\bibinfo}[2]{#2}
\providecommand{\BIBentrySTDinterwordspacing}{\spaceskip=0pt\relax}
\providecommand{\BIBentryALTinterwordstretchfactor}{4}
\providecommand{\BIBentryALTinterwordspacing}{\spaceskip=\fontdimen2\font plus
\BIBentryALTinterwordstretchfactor\fontdimen3\font minus
  \fontdimen4\font\relax}
\providecommand{\BIBforeignlanguage}[2]{{%
\expandafter\ifx\csname l@#1\endcsname\relax
\typeout{** WARNING: IEEEtranN.bst: No hyphenation pattern has been}%
\typeout{** loaded for the language `#1'. Using the pattern for}%
\typeout{** the default language instead.}%
\else
\language=\csname l@#1\endcsname
\fi
#2}}
\providecommand{\BIBdecl}{\relax}
\BIBdecl

\bibitem[Patel et~al.(2018)Patel, Emery, and Chen]{patel2018contextualnet}
M.~Patel, B.~Emery, and Y.-Y. Chen, ``{ContextualNet}: Exploiting contextual
  information using {LSTMs} to improve image-based localization,'' in
  \emph{Proc. {IEEE} Int. Conf. Robot. and Automation}.\hskip 1em plus 0.5em
  minus 0.4em\relax IEEE, 2018, pp. 1--7.

\bibitem[Mur-Artal and Tard{\'o}s(2017)]{mur2017orb}
R.~Mur-Artal and J.~D. Tard{\'o}s, ``{ORB-SLAM}2: An open-source {SLAM} system
  for monocular, stereo, and {RGB-D} cameras,'' \emph{{IEEE} Trans. Robot.},
  vol.~33, no.~5, pp. 1255--1262, 2017.

\bibitem[Kendall et~al.(2015)Kendall, Grimes, and Cipolla]{kendall2015posenet}
A.~Kendall, M.~Grimes, and R.~Cipolla, ``{PoseNet}: A convolutional network for
  real-time 6-{DOF} camera relocalization,'' in \emph{Proc. {IEEE} Int. Conf.
  Comput. Vis.}, 2015, pp. 2938--2946.

\bibitem[Shotton et~al.(2013)Shotton, Glocker, Zach, Izadi, Criminisi, and
  Fitzgibbon]{shotton2013scene}
J.~Shotton, B.~Glocker, C.~Zach, S.~Izadi, A.~Criminisi, and A.~Fitzgibbon,
  ``Scene coordinate regression forests for camera relocalization in {RGB-D}
  images,'' in \emph{Proc. {IEEE} Conf. Comput. Vis. Pattern Recog.}, 2013, pp.
  2930--2937.

\bibitem[Wang et~al.(2019)Wang, Chen, Lu, Zhao, Trigoni, and
  Markham]{wang2019atloc}
B.~Wang, C.~Chen, C.~X. Lu, P.~Zhao, N.~Trigoni, and A.~Markham, ``{AtLoc}:
  Attention guided camera localization,'' \emph{arXiv preprint
  arXiv:1909.03557}, 2019.

\bibitem[Weinzaepfel et~al.(2019)Weinzaepfel, Csurka, Cabon, and
  Humenberger]{weinzaepfel2019visual}
P.~Weinzaepfel, G.~Csurka, Y.~Cabon, and M.~Humenberger, ``Visual localization
  by learning objects-of-interest dense match regression,'' in \emph{Proc.
  {IEEE} Conf. Comput. Vis. Pattern Recog.}, 2019, pp. 5634--5643.

\bibitem[Valada et~al.(2018)Valada, Radwan, and Burgard]{valada2018deep}
A.~Valada, N.~Radwan, and W.~Burgard, ``Deep auxiliary learning for visual
  localization and odometry,'' in \emph{Proc. {IEEE} Int. Conf. Robot. and
  Automation}.\hskip 1em plus 0.5em minus 0.4em\relax IEEE, 2018, pp.
  6939--6946.

\bibitem[Radwan et~al.(2018)Radwan, Valada, and Burgard]{radwan2018vlocnet++}
N.~Radwan, A.~Valada, and W.~Burgard, ``{VlocNet++}: Deep multitask learning
  for semantic visual localization and odometry,'' \emph{{IEEE} Robot. and
  Autom. Lett.}, vol.~3, no.~4, pp. 4407--4414, 2018.

\bibitem[Tian et~al.(2020)Tian, Nie, and Shen]{tian20203d}
M.~Tian, Q.~Nie, and H.~Shen, ``{3D} scene geometry-aware constraint for camera
  localization with deep learning,'' in \emph{Proc. {IEEE} Int. Conf. Robot.
  and Automation}.\hskip 1em plus 0.5em minus 0.4em\relax IEEE, 2020, pp.
  4211--4217.

\bibitem[Kendall and Cipolla(2016)]{kendall2016modelling}
A.~Kendall and R.~Cipolla, ``Modelling uncertainty in deep learning for camera
  relocalization,'' in \emph{Proc. {IEEE} Int. Conf. Robot. and
  Automation}.\hskip 1em plus 0.5em minus 0.4em\relax IEEE, 2016, pp.
  4762--4769.

\bibitem[Zhao et~al.(2019)Zhao, Shen, Sun, and Yang]{zhao2019generative}
C.~Zhao, M.~Shen, L.~Sun, and G.-Z. Yang, ``Generative localization with
  uncertainty estimation through {Video-CT} data for bronchoscopic biopsy,''
  \emph{{IEEE} Robot. and Autom. Lett.}, vol.~5, no.~1, pp. 258--265, 2019.

\bibitem[Costante and Mancini(2020)]{costante2020uncertainty}
G.~Costante and M.~Mancini, ``Uncertainty estimation for data-driven visual
  odometry,'' \emph{{IEEE} Trans. Robot.}, vol.~36, no.~6, pp. 1738--1757,
  2020.

\bibitem[Yang et~al.(2020)Yang, Stumberg, Wang, and Cremers]{yang2020d3vo}
N.~Yang, L.~v. Stumberg, R.~Wang, and D.~Cremers, ``{D3VO}: Deep depth, deep
  pose and deep uncertainty for monocular visual odometry,'' in \emph{Proc.
  {IEEE} Conf. Comput. Vis. Pattern Recog.}, 2020, pp. 1281--1292.

\bibitem[Liu et~al.(2018)Liu, Ok, Vega-Brown, and Roy]{liu2018deep}
K.~Liu, K.~Ok, W.~Vega-Brown, and N.~Roy, ``Deep inference for covariance
  estimation: Learning {Gussian} noise models for state estimation,'' in
  \emph{Proc. {IEEE} Int. Conf. Robot. and Automation}.\hskip 1em plus 0.5em
  minus 0.4em\relax IEEE, 2018, pp. 1436--1443.

\bibitem[Hartley and Zisserman(2003)]{hartley2003multiple}
R.~Hartley and A.~Zisserman, \emph{Multiple view geometry in computer
  vision}.\hskip 1em plus 0.5em minus 0.4em\relax Cambridge university press,
  2003.

\bibitem[Vega-Brown and Roy(2013)]{vega2013cello}
W.~Vega-Brown and N.~Roy, ``{CELLO-EM}: Adaptive sensor models without ground
  truth,'' in \emph{Proc. {IEEE}/{RSJ} Int. Conf. Intell. Robots and
  Syst.}\hskip 1em plus 0.5em minus 0.4em\relax IEEE, 2013, pp. 1907--1914.

\bibitem[Ozog and Eustice(2013)]{ozog2013importance}
P.~Ozog and R.~M. Eustice, ``On the importance of modeling camera calibration
  uncertainty in visual {SLAM},'' in \emph{Proc. {IEEE} Int. Conf. Robot. and
  Automation}.\hskip 1em plus 0.5em minus 0.4em\relax IEEE, 2013, pp.
  3777--3784.

\bibitem[Polic and Pajdla(2017)]{polic2017camera}
M.~Polic and T.~Pajdla, ``Camera uncertainty computation in large {3D}
  reconstruction,'' in \emph{Int. Conf. on 3D Vis.}\hskip 1em plus 0.5em minus
  0.4em\relax IEEE, 2017, pp. 282--290.

\bibitem[Briskin et~al.(2017)Briskin, Geva, Rivlin, and
  Rotstein]{briskin2017estimating}
G.~Briskin, A.~Geva, E.~Rivlin, and H.~Rotstein, ``Estimating pose and motion
  using bundle adjustment and digital elevation model constraints,''
  \emph{{IEEE} Trans. Aerosp. Electron. Syst.}, vol.~53, no.~4, pp. 1614--1624,
  2017.

\bibitem[G{\'a}lvez-L{\'o}pez and Tardos(2012)]{galvez2012bags}
D.~G{\'a}lvez-L{\'o}pez and J.~D. Tardos, ``Bags of binary words for fast place
  recognition in image sequences,'' \emph{{IEEE} Trans. Robot.}, vol.~28,
  no.~5, pp. 1188--1197, 2012.

\bibitem[Cummins and Newman(2008)]{cummins2008fab}
M.~Cummins and P.~Newman, ``{FAB-MAP}: Probabilistic localization and mapping
  in the space of appearance,'' \emph{Int. J. Robot. Res.}, vol.~27, no.~6, pp.
  647--665, 2008.

\bibitem[S{\"u}nderhauf et~al.(2015)S{\"u}nderhauf, Shirazi, Dayoub, Upcroft,
  and Milford]{sunderhauf2015performance}
N.~S{\"u}nderhauf, S.~Shirazi, F.~Dayoub, B.~Upcroft, and M.~Milford, ``On the
  performance of convnet features for place recognition,'' in \emph{Proc.
  {IEEE}/{RSJ} Int. Conf. Intell. Robots and Syst.}\hskip 1em plus 0.5em minus
  0.4em\relax IEEE, 2015, pp. 4297--4304.

\bibitem[Kim et~al.(2017)Kim, Dunn, and Frahm]{kim2017learned}
H.~J. Kim, E.~Dunn, and J.-M. Frahm, ``Learned contextual feature reweighting
  for image geo-localization,'' in \emph{Proc. {IEEE} Conf. Comput. Vis.
  Pattern Recog.}\hskip 1em plus 0.5em minus 0.4em\relax IEEE, 2017, pp.
  3251--3260.

\bibitem[Brahmbhatt et~al.(2018)Brahmbhatt, Gu, Kim, Hays, and
  Kautz]{brahmbhatt2018geometry}
S.~Brahmbhatt, J.~Gu, K.~Kim, J.~Hays, and J.~Kautz, ``Geometry-aware learning
  of maps for camera localization,'' in \emph{Proc. {IEEE} Conf. Comput. Vis.
  Pattern Recog.}, 2018, pp. 2616--2625.

\bibitem[Zagoruyko and Komodakis(2015)]{zagoruyko2015learning}
S.~Zagoruyko and N.~Komodakis, ``Learning to compare image patches via
  convolutional neural networks,'' in \emph{Proc. {IEEE} Conf. Comput. Vis.
  Pattern Recog.}, 2015, pp. 4353--4361.

\bibitem[G\'alvez-L\'opez and Tard\'os(2012)]{GalvezTRO12}
D.~G\'alvez-L\'opez and J.~D. Tard\'os, ``Bags of binary words for fast place
  recognition in image sequences,'' \emph{{IEEE} Trans. Robot.}, vol.~28,
  no.~5, pp. 1188--1197, October 2012.

\bibitem[Schmitt and Schuller(2017)]{schmitt2017openxbow}
M.~Schmitt and B.~Schuller, ``{OpenXBOW}: Introducing the passau open-source
  crossmodal bag-of-words toolkit,'' \emph{J. Mach. Learning Res.}, vol.~18,
  no.~1, pp. 3370--3374, 2017.

\bibitem[Wang et~al.(2014)Wang, Song, Leung, Rosenberg, Wang, Philbin, Chen,
  and Wu]{wang2014learning}
J.~Wang, Y.~Song, T.~Leung, C.~Rosenberg, J.~Wang, J.~Philbin, B.~Chen, and
  Y.~Wu, ``Learning fine-grained image similarity with deep ranking,'' in
  \emph{Proc. {IEEE} Conf. Comput. Vis. Pattern Recog.}, 2014, pp. 1386--1393.

\bibitem[Szegedy et~al.(2015)Szegedy, Liu, Jia, Sermanet, Reed, Anguelov,
  Erhan, Vanhoucke, and Rabinovich]{szegedy2015going}
C.~Szegedy, W.~Liu, Y.~Jia, P.~Sermanet, S.~Reed, D.~Anguelov, D.~Erhan,
  V.~Vanhoucke, and A.~Rabinovich, ``Going deeper with convolutions,'' in
  \emph{Proc. {IEEE} Conf. Comput. Vis. Pattern Recog.}, 2015, pp. 1--9.

\bibitem[Hadsell et~al.(2006)Hadsell, Chopra, and
  LeCun]{hadsell2006dimensionality}
R.~Hadsell, S.~Chopra, and Y.~LeCun, ``Dimensionality reduction by learning an
  invariant mapping,'' in \emph{Proc. {IEEE} Conf. Comput. Vis. Pattern
  Recog.}, vol.~2.\hskip 1em plus 0.5em minus 0.4em\relax IEEE, 2006, pp.
  1735--1742.

\bibitem[Ondr{\'u}{\v{s}}ka et~al.(2015)Ondr{\'u}{\v{s}}ka, Kohli, and
  Izadi]{ondruvska2015mobilefusion}
P.~Ondr{\'u}{\v{s}}ka, P.~Kohli, and S.~Izadi, ``Mobilefusion: Real-time
  volumetric surface reconstruction and dense tracking on mobile phones,''
  \emph{{IEEE} Trans. Vis. Comput. Graphics}, vol.~21, no.~11, pp. 1251--1258,
  2015.

\bibitem[Huber(1992)]{huber1992robust}
P.~J. Huber, ``Robust estimation of a location parameter,'' in
  \emph{Breakthroughs in statistics}.\hskip 1em plus 0.5em minus 0.4em\relax
  Springer, 1992, pp. 492--518.

\bibitem[K{\"u}mmerle et~al.(2011)K{\"u}mmerle, Grisetti, Strasdat, Konolige,
  and Burgard]{kummerle2011g}
R.~K{\"u}mmerle, G.~Grisetti, H.~Strasdat, K.~Konolige, and W.~Burgard, ``{G} 2
  o: A general framework for graph optimization,'' in \emph{Proc. {IEEE} Int.
  Conf. Robot. and Automation}.\hskip 1em plus 0.5em minus 0.4em\relax IEEE,
  2011, pp. 3607--3613.

\bibitem[Barfoot(2017)]{barfoot2017state}
T.~D. Barfoot, \emph{State estimation for robotics}.\hskip 1em plus 0.5em minus
  0.4em\relax Cambridge University Press, 2017.

\bibitem[Thrun(2002)]{thrun2002probabilistic}
S.~Thrun, ``Probabilistic robotics,'' \emph{Communications of the ACM},
  vol.~45, no.~3, pp. 52--57, 2002.

\bibitem[Shi et~al.(2002)Shi, Chan, Wong, and Ho]{shi2002speed}
K.~Shi, T.~Chan, Y.~Wong, and S.~Ho, ``Speed estimation of an induction motor
  drive using an optimized extended {Kalman} filter,'' \emph{IEEE Trans. on
  Indust. Electron.}, vol.~49, no.~1, pp. 124--133, 2002.

\bibitem[Barfoot and Furgale(2014)]{barfoot2014associating}
T.~D. Barfoot and P.~T. Furgale, ``Associating uncertainty with
  three-dimensional poses for use in estimation problems,'' \emph{{IEEE} Trans.
  Robot.}, vol.~30, no.~3, pp. 679--693, 2014.

\bibitem[Andrews(2021)]{issacsdk}
\BIBentryALTinterwordspacing
G.~Andrews, ``{NVIDIA} isaac sim on omniverse now available in open beta,'' Jun
  2021. [Online]. Available:
  \url{https://developer.nvidia.com/blog/nvidia-isaac-sim-on-omniverse-now-available-in-open-beta/}
\BIBentrySTDinterwordspacing

\bibitem[Sturm et~al.(2012)Sturm, Engelhard, Endres, Burgard, and
  Cremers]{sturm12iros}
J.~Sturm, N.~Engelhard, F.~Endres, W.~Burgard, and D.~Cremers, ``A benchmark
  for the evaluation of {RGB-D} {SLAM} systems,'' in \emph{Proc. {IEEE}/{RSJ}
  Int. Conf. Intell. Robots and Syst.}, Oct. 2012.

\bibitem[Newcombe et~al.(2011)Newcombe, Izadi, Hilliges, Molyneaux, Kim,
  Davison, Kohi, Shotton, Hodges, and Fitzgibbon]{newcombe2011kinectfusion}
R.~A. Newcombe, S.~Izadi, O.~Hilliges, D.~Molyneaux, D.~Kim, A.~J. Davison,
  P.~Kohi, J.~Shotton, S.~Hodges, and A.~Fitzgibbon, ``{KinectFusion}:
  Real-time dense surface mapping and tracking,'' in \emph{{IEEE} Internat.
  Sympo. on Mixed and Augment. Reality}.\hskip 1em plus 0.5em minus 0.4em\relax
  IEEE, 2011, pp. 127--136.

\bibitem[Kingma and Ba(2014)]{kingma:2014}
D.~P. Kingma and J.~Ba, ``{Adam}: {A} method for stochastic optimization,''
  \emph{Intern. Conf. for Learning Represent.}, 2014.

\bibitem[Abadi et~al.(2016)Abadi, Agarwal, Barham, Brevdo, Chen, Citro,
  Corrado, Davis, Dean, Devin, et~al.]{abadi2016tensorflow}
M.~Abadi, A.~Agarwal, P.~Barham, E.~Brevdo, Z.~Chen, C.~Citro, G.~S. Corrado,
  A.~Davis, J.~Dean, M.~Devin \emph{et~al.}, ``Tensorflow: Large-scale machine
  learning on heterogeneous distributed systems,'' \emph{arXiv preprint
  arXiv:1603.04467}, 2016.

\bibitem[Zhou et~al.(2014)Zhou, Lapedriza, Xiao, Torralba, and
  Oliva]{zhou:2014}
B.~Zhou, A.~Lapedriza, J.~Xiao, A.~Torralba, and A.~Oliva, ``Learning deep
  features for scene recognition using places database,'' in \emph{Proc.
  Advances Neural Inform. Process. Syst. Conf.}, 2014, pp. 487--495.

\end{thebibliography}
	}
	
\end{document}